\documentclass[letterpaper, 10 pt, conference]{ieeeconf}  

\IEEEoverridecommandlockouts                              

\usepackage{latexsym}
\usepackage{amsmath}
\usepackage{amssymb}
\usepackage{epsfig}
\usepackage{bbm}
\usepackage[tight]{subfigure}

\usepackage{cite}

\usepackage[ruled]{algorithm2e}

\usepackage{verbatim}
\usepackage{hyperref}
\hypersetup{%
  breaklinks,
  bookmarksnumbered=true,
  bookmarksopen=true,
  colorlinks=true,
  linkcolor=black,
  urlcolor=black,
  citecolor=black
}
\usepackage{xcolor}
\usepackage{subfiles}

\begin{document}
\bstctlcite{IEEEexample:BSTcontrol}
%
\title{TartanDrive: A Large-Scale Dataset for Learning\\ Off-Road Dynamics Models}




\author{Samuel Triest$^{1}$, Matthew Sivaprakasam$^{2}$, Sean J. Wang$^{3}$,\\ Wenshan Wang$^{1}$, Aaron M. Johnson$^{3}$, and Sebastian Scherer$^{1}$
\thanks{* This work was supported by ARL awards \#W911NF1820218 and \#W911NF20S0005.}%
\thanks{$^{1}$ Robotics Institute, Carnegie Mellon University, Pittsburgh, PA, USA. \{striest,wenshanw,basti\}@andrew.cmu.edu}%
\thanks{$^{2}$ Electrical \& Computer Engineering Dept., University of Pittsburgh, Pittsburgh, PA, USA. mjs299@pitt.edu}%
\thanks{$^{3}$ Mechanical Engineering Dept., Carnegie Mellon University, Pittsburgh, PA, USA. \{sjw2, amj1\}@andrew.cmu.edu}%
}

\maketitle

%
\IEEEpeerreviewmaketitle

\begin{abstract}

We present TartanDrive, a large scale dataset for learning dynamics models for off-road driving. We collected a dataset of roughly 200,000 off-road driving interactions on a modified Yamaha Viking ATV with seven unique sensing modalities in diverse terrains. To the authors' knowledge, this is the largest real-world multi-modal off-road driving dataset, both in terms of number of interactions and sensing modalities. We also benchmark several state-of-the-art methods for model-based reinforcement learning from high-dimensional observations on this dataset. We find that extending these models to multi-modality leads to significant performance on off-road dynamics prediction, especially in more challenging terrains. We also identify some shortcomings with current neural network architectures for the off-road driving task. Our dataset is available at \href{https://github.com/castacks/tartan_drive}{https://github.com/castacks/tartan\_drive}.


\end{abstract}


\section{Introduction}










Robots need to understand the physical properties of the world to deal with its complexity in practical tasks such as off-road driving. Robots should not only rely on the geometric and semantic information of observed scenes, but also reason about dynamical features, such as the vehicle speed and approach angle, to avoid getting stuck or damaged in various types of terrain, such as puddles, tall grass, and loose gravel.  

Modeling this complex interplay between the robot and the environment is extremely difficult using traditional methods, which typically rely on high-fidelity models of both the dynamics of the robot, as well as the environment that it interacts with \cite{howard2007optimal}. Data-driven methods \cite{kahn2021badgr, tremblay2020multimodal, divconstrained}  have been explored to address this issue and have shown great potential, but still require large-scale diverse datasets for the training.

\begin{figure}
    \centering
    \includegraphics[width=\linewidth]{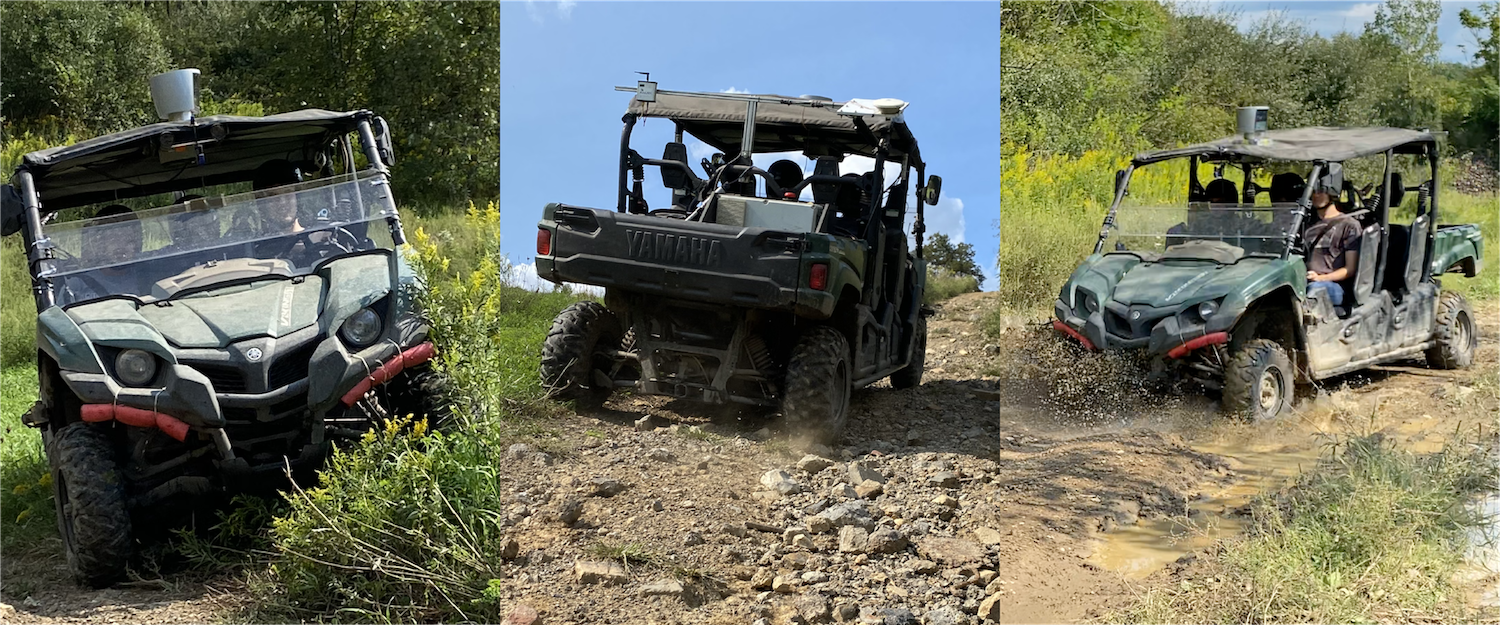}
    \caption{We provide a rich dataset for learning off-road vehicle dynamics. Data was collected by driving a modified ATV through a variety of terrain including tall grass, rocks, and mud. Driving data consists of robot actions, and a variety of multi-modal observations.}
    \label{fig:atv_terrain}
\end{figure}

\begin{figure*}
    \centering
    \includegraphics[width=\linewidth, trim={0 2cm 0 1cm }]{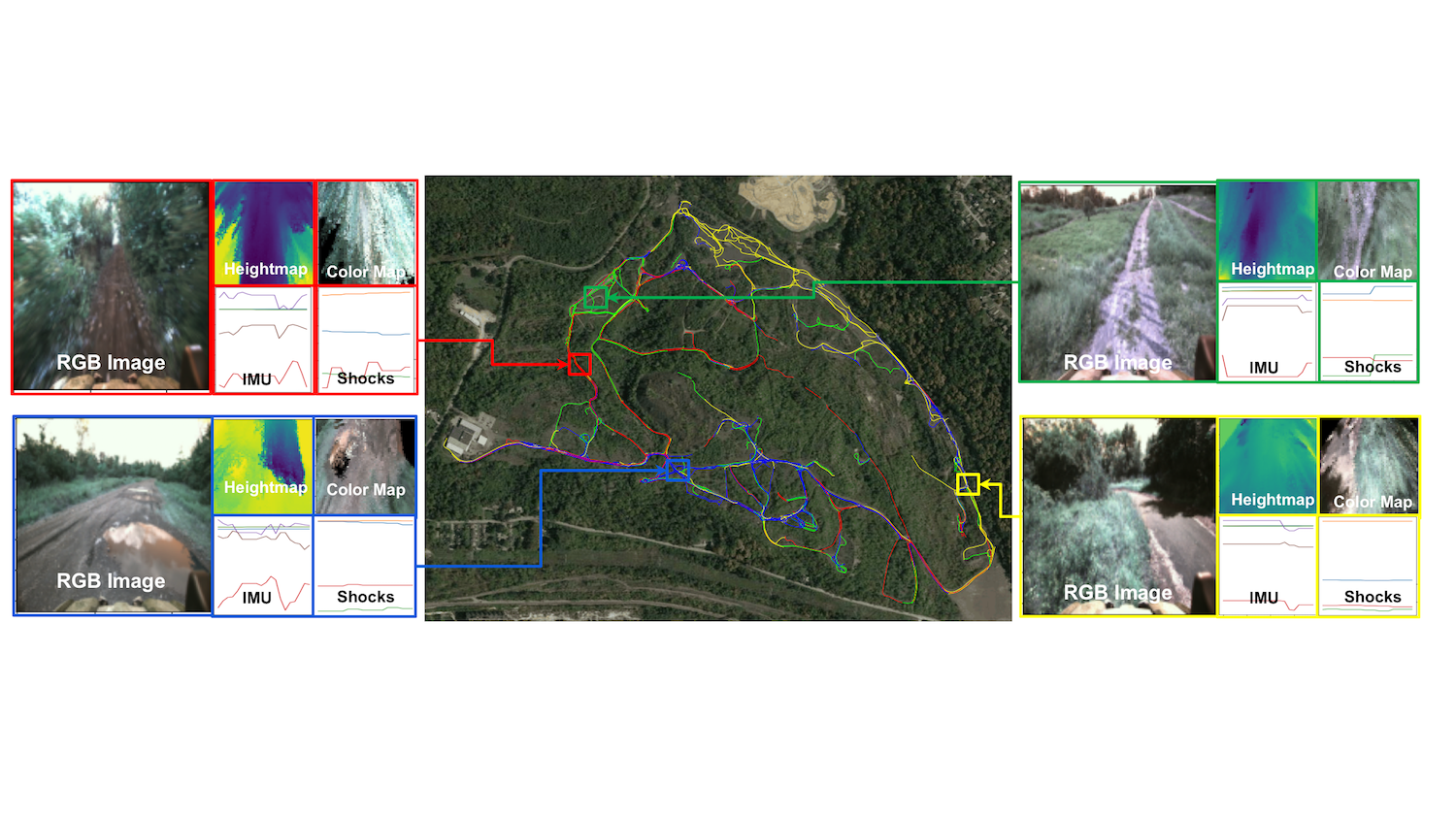}
    \caption{Our dataset contains several hours of driving data from a test site in Pittsburgh, PA. Our dataset contains diverse off-road driving scenarios and several sensing modalities, including front-facing camera and top-down local maps. Shown in the center is a satellite image of the testing site with trajectories superimposed. They are colored according to a clustering based on ResNet \cite{he2016identity} features on the RGB images. Shown on the sides are four datapoints (one from each cluster) with five of the seven modalities available in the dataset shown. We can observe a diverse set of scenarios, including dense foliage (red), steep slopes (green) open road (blue) and puddles (yellow).}
    \label{fig:title_fig}
\end{figure*}

While many autonomous driving datasets exist, most focus on urban environments \cite{geyer2020a2d2,waymo_open_dataset,Argoverse,Cordts2016Cityscapes,RobotCarDatasetIJRR}. For off-road driving, existing datasets only focus on scene understanding, with a special focus on semantic segmentation \cite{maturana2018real,RUGD2019IROS,valada2016deep,dabbiru2020lidar,jiang2020rellis3d, shaban2022semantic}. We argue the semantic labels (i.e.\ dirt, grass, tree, bush, mud) are not sufficient. For example, a bush might be traversable for one robot but not for another less capable robot. Additionally, a robot could get stuck in mud at low speed but not at high speed. This information is hard to obtain from camera and LiDAR sensors alone. Rather, it is more feasible to learn these properties from \textit{interaction data}, which can include data from robot actions, wheel encoders and IMU.  

There is also an increasing interest in developing datasets for understanding physical common-sense and predicting interactions between objects. However, they either utilize a simulated world with simple objects such as a pile of blocks \cite{battaglia2016interaction,lerer2016learning,baradel2019cophy} or are collected in controlled laboratory environments \cite{agrawal2016learning,levine2018learning,finn2016unsupervised}. 

We present a large and diverse real-world off-road driving dataset, which contains multi-modal interaction data in various complex terrains, including challenging interactions such as driving through dense vegetation and puddles, driving on steep slopes, and driving at high speed with tire slip. We believe this dataset will not only benefit the development of autonomous off-road driving, but also facilitate research in modeling complex robot dynamics.

This dataset is motivated by two points. First, accurately modeling off-road vehicle dynamics is difficult, especially in real-world scenarios. While some approaches leverage analytical models of the terrain \cite{howard2007optimal}, these models are often low fidelity since it is intractable to correctly model complex interactions such as collisions with rocks and gravel, tire interactions with arbitrary surfaces, etc. Recent work \cite{kahn2021badgr, tremblay2020multimodal} leverages off-road driving datasets to create data-driven dynamics models instead. However, existing datasets are limited in either the amount or diversity of data, or the amount of modalities available. 
Second, data-driven models can benefit from supervision from multi-modal sensory input. 
Tremblay et al. \cite{tremblay2020multimodal,tremblay2019automatic} have shown that leveraging various sensing modalities leads to learned models that are more robust to challenging and inconstant dynamics.

The contributions of this work are as follows:

\begin{enumerate}
    \item A large-scale dataset collected on a Yamaha ATV in off-road scenarios, containing roughly 200,000 interaction datapoints with multiple high-dimensional sensing modalities. To the best of the authors' knowledge, this exceeds the largest multi-modal interaction dataset for off-road driving \cite{tremblay2019automatic} in both the number of available datapoints and modalities.
    
    \item Benchmarks of several state-of-the-art neural network architectures and training procedures \cite{hafner2019dream, hafner2019learning, hafner2020mastering} for high-dimensional observations in a real-world scenario with challenging dynamics.
\end{enumerate}

\section{Related Work}
Most publicly-available off-road driving datasets focus on understanding environmental features instead of the interplay between the robot and the environment \cite{pezzementi2018comparing}. 
RUGD \cite{RUGD2019IROS} consists of video sequences with segmentation label of 24 unique category annotations. 
Maturana et al. \cite{maturana2018real} collected a segmentation dataset, which tried to explore more fine-grained information such as traversable grass and non-traversable vegetation, but those discrete labels are still too abstract to understand how a robot is going to behave in the specific case. 
Gresenz et al. \cite{gresenzoff} collected a dataset containing over 10000 images of offroad bicycle driving and labels corresponding to the roughness of the terrain in the image, where the labels were generated from processing sensory data available on the platform. Similarly, this dataset does not provide enough interaction data or actions. 
Rellis 3D \cite{jiang2020rellis3d} consists of images, pointclouds, robot states, and actions, but the amount of trajectory data is rather small (about 20 minutes -- 12,000 datapoints at 10hz -- total). 
Generally, learning dynamics models requires on the order of hundreds of thousands of dynamics interactions. As our dataset is designed to focus on dynamics models with multi-modal sensory inputs without the need for hand-labeling, we are able to collect much more data than existing datasets.

More and more off-road driving research has taken account of the robot-environment interaction rather than environment features alone to improve the driving performance. Due to the lack of publicly-available real-world datasets, they usually train their models in simulation environments or collect their own data in a small scale or for a specific research purpose. 
Tremblay et al. \cite{tremblay2020multimodal} trained a multi-modal dynamics model in a simulation environment based on Unreal Engine, and a small forest dataset (Montmorency) \cite{tremblay2019automatic}.  
Sivaprakasam et al. \cite{sivaprakasm2021} developed a simulation environment with random obstacles to learn a predictive model from physical interaction data. Similarly, Wang et al. \cite{divconstrained} trained a probabilistic dynamics model for planning using a simulator and a small robot platform. 
Kahn et al. \cite{kahn2021badgr} employ a model-based RL technique in order to enable a wheeled robot to navigate off-road terrain using a monocular front-facing camera \cite{kahn2021badgr}. Their algorithm (BADGR) leverages many hours of trajectory data collected via random exploration in order to train a neural network to model the dynamics of the robot over various types of terrain. In addition to using the position of the robot, the authors also utilize hand-designed events calculated from the on-board IMU and LiDAR as additional supervision for the model. The authors make their data publicly-available, where the data consists of the hand-designed events (such as bumpiness and collision), RGB images, robot states, and actions.

On the other hand, learning a dynamics model has been an active research area in various directions such as model-based reinforcement learning (RL). Most work in RL literature learns the model in simulation environments. For the off-road driving task, \cite{kahn2021badgr} learned from the interaction data a classification network to predict the hand-designed events (i.e.\ collision, smoothness, etc.). Tremblay et al. \cite{tremblay2020multimodal} modify the recurrent state-space model of Hafner et al. \cite{hafner2019learning} to handle multiple modalities and provide a modified training objective. Wang et al. \cite{divconstrained} improve trajectory prediction by incorporating uncertainty estimation and a closed-loop tracker. In this paper, we follow these papers and test several neural network architectures to demonstrate the potential value of the proposed TartanDrive dataset.

\begin{table*}[]
    \centering
    \vspace*{0.05in}
    \begin{tabular}{c||c|c|c|c|c|c|c|c|c|c|c}
         Dataset & Samples & State & Action & Image & Pointcloud & Heightmap & RGBmap & IMU & Wheel RPM & Shocks & Intervention\\
         \hline
         RUGD & ~2700 & No & No & Yes & No & No & No & No & No & No & No\\
         Rellis 3D & ~13800 & Yes & Yes & Yes & Yes & No & No & Yes & No & No & No\\
         Montmorency & ~75000 & Yes & Yes & Yes & Yes & Yes & No & Yes & No & No & No \\
         Ours & 184000 & Yes & Yes & Yes & No & Yes & Yes & Yes & Yes & Yes & Yes
    \end{tabular}
    \caption{Overview and comparison of various off-road driving datasets}
    \label{tab:my_label}
\end{table*}




\section{The Dataset}

Our off-road driving dataset is designed for the purpose of dynamics prediction. As such, it contains a large number of interaction data over a diverse set of terrains and scenarios.

\begin{table}[]
    \centering
    \begin{tabular}{c||c|c|c}
        Modality & Type & Dimension & Train Dimension \\
        \hline
        Robot Pose & Vector & 7 & 7\\
        RGB Image & Image & 2 x 1024 x 512 & 128 x 128 \\
        Heightmap & Image & 500 x 500 & 64 x 64\\
        RGB Map & Image & 500 x 500 & 64 x 64\\
        IMU & Time-series & 20 x 6 & 20 x 6\\
        Shock Position & Time-series & 5 x 4 & 20 x 4 \\
        Wheel RPM & Time-series & 5 x 4 & 20 x 4\\
        Intervention & Boolean & 1 & 1\\
    \end{tabular}
    \caption{The set of available dataset features and their sizes. }
    \label{tab:dataset_modalities}
    
    \vspace*{-0.7cm}
\end{table}




\subsection{ATV Platform}
We use a Yamaha Viking All-Terrain Vehicle (ATV) which was previously modified by Mai et al. \cite{Mai-2020} to collect data. The throttle and steering of the ATV were controlled using a Kairos Autonomi steering ring and servos via a joystick. For safety reasons, the brakes were controlled directly by a human driver.

\begin{figure}
    \centering
    \vspace*{0.07in}
    \includegraphics[width=\linewidth]{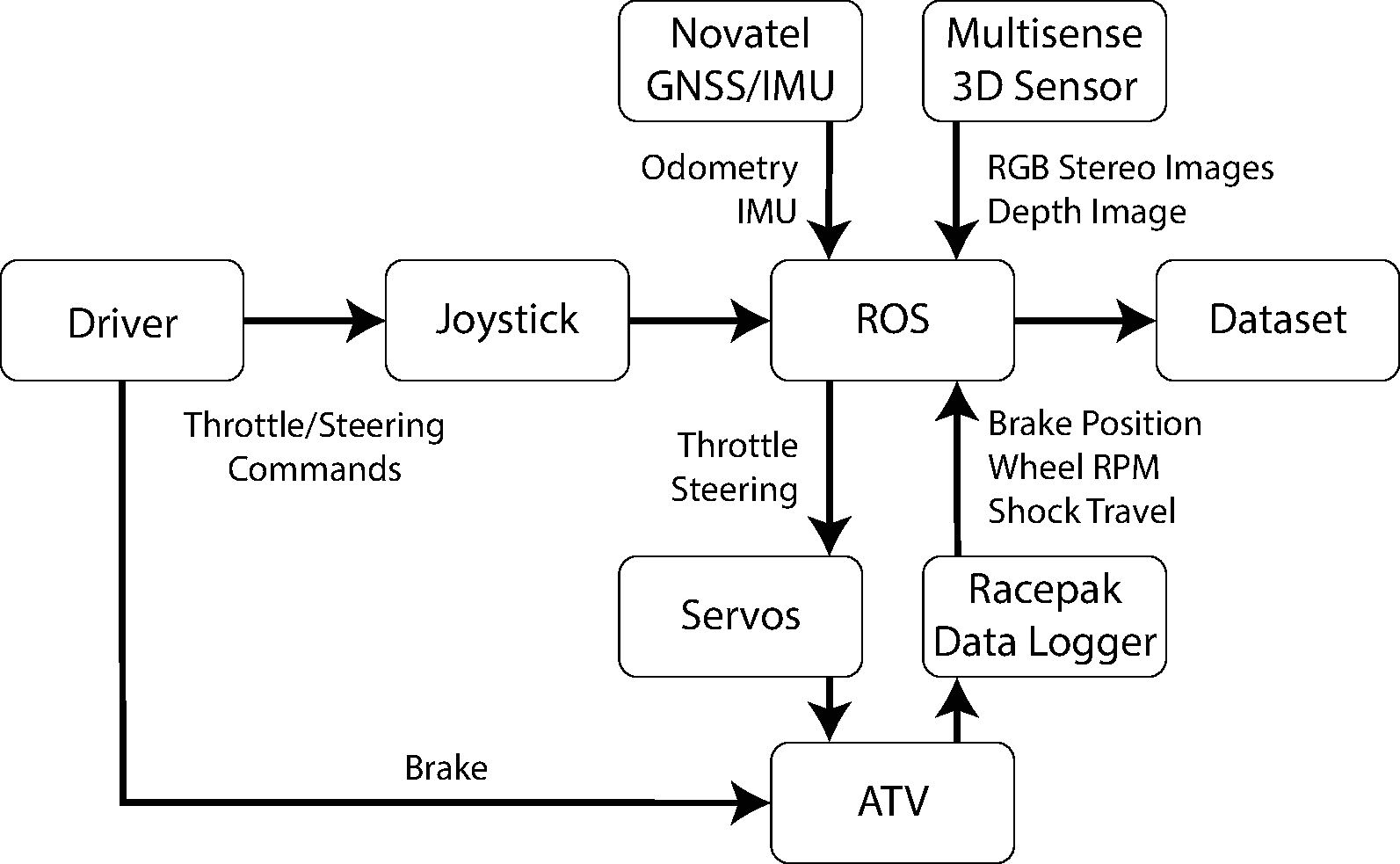}
    \caption{A system diagram of our data collection setup. We use multiple sensors to collect interaction data across many modalities, as well as throttle and steering commands.}
    \label{fig:system_diagram}
\end{figure}

Proprioceptive and exteroceptive sensors were used to collect real time data. First, a forward facing Carnegie Robotics Multisense S21 stereo camera provided long range, high resolution stereo RGB and depth images, as well as inertial measurements. A NovAtel PROPAK-V3-RT2i GNSS unit provided global pose estimates. A Racepak G2X Pro Data Logger recorded suspension shock travel and wheel rpm at all four wheels, as well as the position of the brake pedal. All sensors and servos were connected to an on-board computer running ROS. Joystick control inputs were relayed to the servos. Figure \ref{fig:system_diagram} shows a system diagram.

\subsection{Data Collection}
We collected data in a diverse range of terrains, including rocks, mud, foliage, as shown in Figure \ref{fig:atv_terrain}. During data collection, we opted to use human tele-operated controls instead of random controls since it may have resulted in catastrophic injuries to the vehicle or on-board passengers. In total, we collected 630 trajectories equivalent to five hours of data. Each trajectory was kept short and ended whenever the driver intervened by applying the brakes.

Data were collected at 10Hz, and consisted of raw sensor data as well as post-processed data. Each observation modality is described below with a summary in Table \ref{tab:dataset_modalities}. Note that we upsample the 50Hz wheel proprioception to 200Hz to match the frequency of the IMU.

\subsubsection{Robot Action} Actions $a=(\mu_t,\mu_s)$ were two-dimensional and corresponded to desired throttle and steering positions. Throttle commands took values between 0 and 1, with 1 corresponding to wide open throttle. Steering commands took values between -1 and 1, with -1 corresponding to a hard left turn. The commands were executed by the servos using PID position control.

\subsubsection{Robot Pose} Robot poses were estimated from the NovAtel GNSS unit. They took the form of a concatenated position vector $p=(x,y,z)$ and quaternion orientation $q=(q_x,q_y,q_z,q_w)$.

\subsubsection{Images} At each timestep, two RGB images were recorded from the onboard stereo camera.

\subsubsection{Local Terrain Maps} 
\begin{figure*}
    \centering
    \vspace*{0.05in}
    \includegraphics[width=0.8\linewidth]{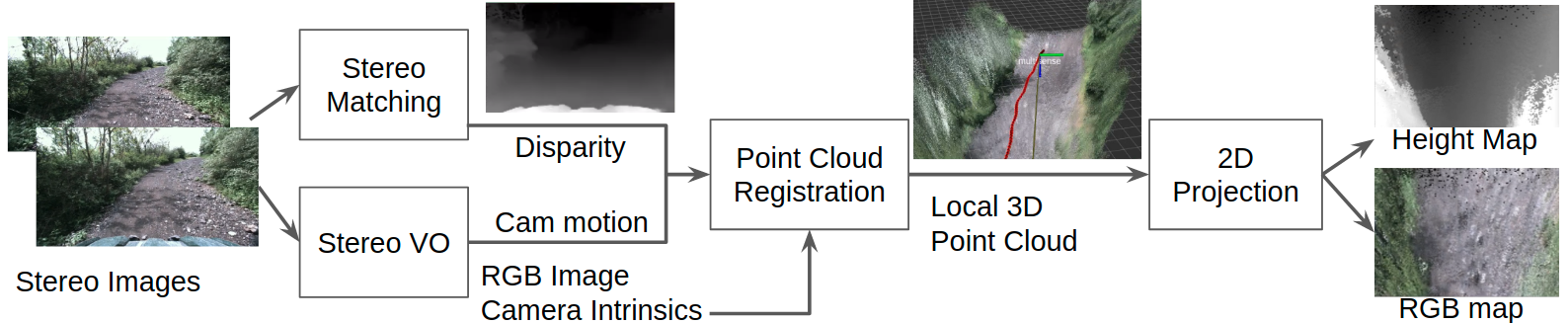}
    \caption{A diagram of the mapping pipeline. We use the stereo images from the Multisense S21 sensor to generate a top-down height map and a top-down RGB map.}
    \label{fig:mapping_diagram}
\end{figure*}

We generate a local top-down-view height map $M_h \in \mathbb{R}^{(w \times h \times 2)}$ (two channels to represent the minimum height and maximum height) and a local RGB map $M_c \in \mathbb{N}^{(w \times h \times 3)}$ using the stereo images from the Multisense S21 sensor. As shown in Fig. \ref{fig:mapping_diagram}, we first generate a disparity image using a stereo matching network \cite{chang2018pyramid}, and estimate the camera motion using TartanVO network \cite{wang2020tartanvo}. Given the camera intrinsics and the corresponding RGB image, we register and colorize each pointcloud into the camera's current frame. Then the registered local 3D point cloud is projected to the ground plane, and then binned, producing a top-down map. We used a resolution of $0.02$ m/pixel, and a region of $(0, 10)$ m in the forward direction, $(-5,5)$ m in the lateral direction, which result in a map size of $500 \times 500$. The maps are updated at 10 Hz.

\subsubsection{Proprioceptive Data} We recorded inertial data (angular velocity and linear acceleration) from the Multisense sensor, as well as shock travel and wheel rpm from the Racepak data logger. Since these sensors made measurements faster than 10Hz, we recorded multiple measurements at each time step in the form of time-series data. The inertial data had a time-series length of 20. The shock travel and wheel rpm data had a time-series length of 5.

\subsubsection{Intervention Data}
Due to safety concerns, the driver was allowed to stop the ATV using the brake pedal. We recorded the brake pedal position, and a boolean intervention signal that indicated when the brakes exceeded a threshold.

\subsection{Dynamical Variation in the Dataset}
How the robot moves depends on the physical properties of the robot, the action command we send, and the physical properties of the environment. In simple environments such as urban roads, the robot properties and actions are sufficient to perform accurate trajectory prediction. We thus ask the question: does our dataset capture the dynamical variation induced by different types of terrains? To answer this question, we first perform a motivational experiment to quantify the correlation between future states and action sequences. The rationale behind this experiment is that if significant dynamical variations exist due to different terrains, then we would expect that similar sequences of actions in the dataset may yield very different trajectories. 

\begin{figure}
    \centering
    \includegraphics[scale=0.45, trim={0 3cm 8cm 0 }]{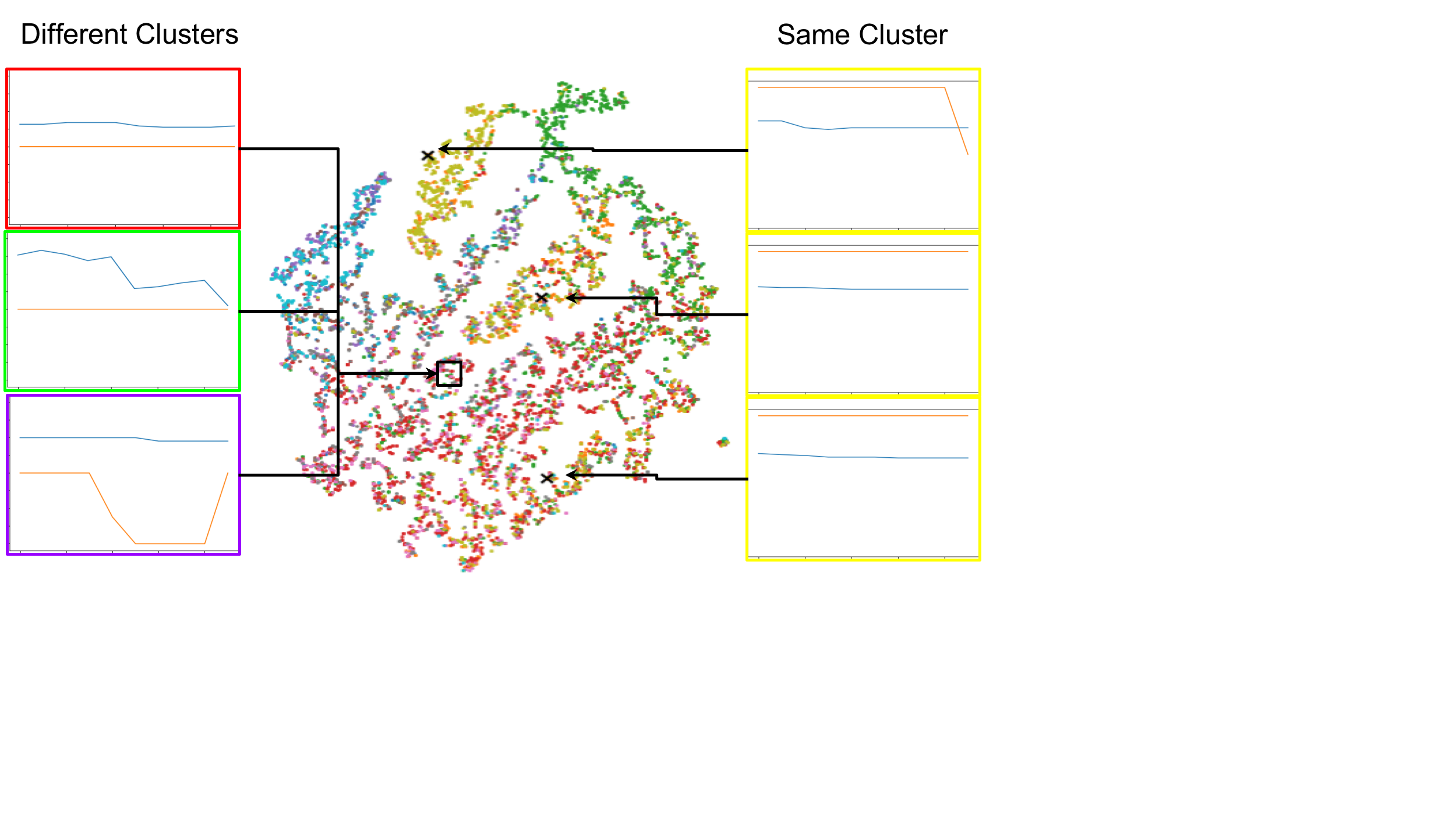}
    \caption{One t-SNE embedding of the state displacements in the dataset. Each point is colored according to its closest action sequence centroid. We can observe from this figure that there is some correlation between the clusters and their position in the t-SNE embedding, though the colors clearly mix. Shown on the left are three action sequences from different clusters that map to the same region in the embedding space. Conversely, shown on the right are three action sequences that map to very different regions of the embedding space, despite being very similar.}
    \label{fig:state_action_tsne}
\end{figure}

In order to perform this experiment, we first collected 10000 random subsequences of length 10 (one second) from our dataset and computed the displacement and rotation from the initial state to the final state (such that every trajectory began from the same initial state). We then performed time-series clustering (using \cite{JMLR:v21:20-091}) on the corresponding sequences of actions. One important note is that we chose simple Euclidean distance instead of time-warping methods \cite{muller2007dynamic} as both the duration and temporal position of the actions in the sequence (and not just the shape of the sequence) affect state displacement. We then computed a t-Distributed Stochastic Neighbor Embedding (t-SNE) \cite{van2008visualizing} of the state displacements and colored each embedded point according to its corresponding action cluster. To mitigate the effect of velocity on the final state displacement, we binned our data based on initial speed and created a separate visualization for each bin. One of the resulting visualizations is shown in Figure \ref{fig:state_action_tsne}. The remainder of the figures, a more detailed desription of the clustering process and the experiment hyperparameters are in the \href{https://github.com/castacks/tartan_drive/blob/main/appendix.pdf}{\color{blue} Appendix} \footnote{ https://github.com/castacks/tartan\_drive/blob/main/appendix.pdf}.



As we can observe in Figure \ref{fig:state_action_tsne}, there is indeed some clustering of state displacements, but many clusters blend together in the visualization. Additionally, the same action cluster does not necessarily form a single cluster in the t-SNE visualization, nor does a single cluster in the t-SNE visualization consist of points belonging to the same action cluster. This confirms our hypothesis that while there is (obviously) a correlation between actions and state displacements, correct dynamics prediction in our dataset requires more features than just the action sequence.

\section{Multi-modal Modeling}

In addition to data collection, we benchmarked several recent neural network architectures for dynamics prediction from high-dimensional data. For this work, we consider the task of dynamics prediction to be the prediction of future states $s_{1:T}$, given an intital state $s_0$, a sequence of actions $a_{1:T}$, and a set of observations $O_0 = \{o^m_0\}$ from a set of modalities $M$. These models thus take the form $f_\theta(s_0, a_{1:T}, O_0)$.




\subsection{Multi-modal Modeling on the ATV}
We first describe the general architecture of our latent-space model for off-road dynamics prediction. Similar to prior work \cite{hafner2019dream, tremblay2020multimodal}, we use a latent-space model that is comprised of three parts:

\begin{enumerate}
    \item An encoder $e_\theta(o_t): \mathcal{O} \rightarrow \mathcal{Z}$ that maps a high-dimensional observation $o_t$ into the latent space
    \item A model $f_\theta(z_t, a_t): (\mathcal{Z},\mathcal{A}) \rightarrow \mathcal{Z}$ that forward-simulates the model in latent space, given actions.
    \item A decoder $d_\theta(z_t): \mathcal{Z} \rightarrow \mathcal{O}$ that maps the low-dimensional latent state back to the observation space.
\end{enumerate}

Parameterizing the model in this fashion allows for efficient state prediction, as one only needs to encode the initial observation $o_0$ to get an initial latent state $z_0$. From this, state vectors $s_{1:T}$ can be recovered via forward-simulating the latent model to get $z_{1:T}$ and then decoding state without decoding the high-dimensional observations. A detailed description of the network architecture is provided in the Appendix. We now describe the specific implementation of our model for the ATV.

\subsubsection*{Multi-modal Encoders}
The encoder for our model consists of a deep neural network for each modality $m$. The original size, and rescaled training size of each modality used is shown in Table \ref{tab:dataset_modalities}. The neural network architectures used for the different modality types are as follows:

\begin{enumerate}
    \item Vector inputs were passed through a dense network to produce a Gaussian distribution $p(z)$.
    \item Images were passed through a convolutional neural network (CNN), flattened, and passed through a dense network to produce $p(z)$.
    \item Time-series inputs were passed through a WaveNet encoder \cite{oord2016wavenet}, flattened, and passed through a dense network.
\end{enumerate}

To combine the multiple predictions on $p(z)$, we follow previous work \cite{wu2018multimodal, tremblay2020multimodal} which use a product of experts \cite{hinton2002training}. In this formulation, the aggregated probability of the latent state, $p(z)$, is determined by the product of probabilities of each expert $\prod_i q_i(z|o_i)$. This formulation is preferable over a mixture of experts for its ability to produce sharper distributions and allow experts to focus on smaller regions of the prediction space.
Since our encoders output diagonal Gaussian distributions in $\mathcal{Z}$, use the result from Cao et al. \cite{cao2014generalized},
%
%
%

\vspace*{-0.05in}
\begin{equation}
\label{eq:poe}
    \begin{split}
         p(z|\textbf{O}) = \prod_i q_i(z|o_i)  = \mathcal{N}\left(\frac{\sum_i(\frac{\mu_i}{\sigma_i})}{\sum_i \frac{1}{\sigma_i}}, \mathbb{I} \left(\sum_i \frac{1}{\sigma_i}\right)\right)
    \end{split}
\end{equation}



\subsubsection*{Latent Model}
Our latent model is implemented as a Gated Recurrent Unit (GRU) \cite{cho2014learning}. While prior work \cite{hafner2019dream, tremblay2020multimodal} use the Recurrent State-Space model, we found that its performance was similar to a GRU for our particular task.

\subsubsection*{Decoders}
We used several different decoder architectures to address the multi-modality of our observations.

\begin{enumerate}
    \item Vector outputs were handled using a dense network.
    \item 
    Image outputs were handled by using deconvolutional layers to upsample $z$.
    \item Time-series outputs were handled by using temporal deconvolutional layers to upsample $z$.
\end{enumerate}

\subsection{Training The Latent Space Model}
We experimented with three different variations of training loss for our experiments. 

\subsubsection{State Reconstruction}
This loss trained the model to maximize log-probability of ground-truth states (position and orientation) from the Novatel, given the initial states, initial observations and sequences of actions. Note that this loss does not train observation decoders.

\vspace*{-0.05in}
\begin{equation}
    \mathcal{L}_{state} = -log p_\theta(s_{1:T}|o_0, s_0, a_{1:T})
\end{equation}

\subsubsection{Reconstruction Loss}
This loss maximizes the log-probability of all observations in addition to the state.

\vspace*{-0.05in}
\begin{equation}
    \mathcal{L}_{rec} = \mathcal{L}_{state} - \sum_m [\beta_m log p_\theta(o^m_{1:T} | z_{1:T})]
\end{equation}

Note that $\beta_m$ allows us to re-weight the importance of each modality.

\subsubsection{Contrastive Loss}
Hafner et al. \cite{hafner2019dream} observed that Bayes' rule can be applied to the reconstruction terms to derive a contrastive loss. Since the contrastive loss is expressed using the latent code $z$, a potential benefit is the ability to ignore distractors in the observation space that are irrelevant to dynamics prediction, e.g.\ image backgrounds. The contrastive loss is defined as:

\vspace*{-0.05in}
\begin{align}
    \mathcal{L}_{con} =  &\mathcal{L}_{state} - 
    \\
   &\beta \bigg[log p_\theta( z_{1:T} | O_{1:T}) - \sum_{O'_{1:T}} log p_\theta(z_{1:T}|O'_{1:T})\bigg],\nonumber
\end{align}
where the added objective aims to maximize the log-probability of the latent code $z$ given the corresponding observation $O$, while minimizing the log-probability of $z$ given the other observations in the batch $O'$. Note that while the terms of the reconstruction loss can be decomposed into independent probabilities, the contrastive loss cannot. As such, there is a single weighting constant $\beta$.


\section{Experiments and Analysis}
Our experiments aim to answer the following questions:
\begin{enumerate}
    \item Does varying the loss type improve model accuracy?
    \item Does using multi-modal sensory data lead to improved dynamics prediction in challenging environments?
\end{enumerate}

\subsection{Does the Loss Type Matter?}
We find that all three loss functions led to similar model accuracy. We attribute this to the fact that in many cases, the high-dimensional sensory inputs are not necessarily correlated with robot motions as they are in the environments used by Hafner et al. \cite{hafner2019dream}. In these simulated environments, predicting future observations is always possible since environments consist only of the agent and a static background. However, future observations are much more difficult to predict in our scenarios. For example, if the ATV drives around a corner, it will be unable to predict observations without some form of mapping and prior traversal. As such, we observe that the auxiliary task of predicting sensory input yields little performance increase.

\subsection{Does Adding Additional Modalities Help?}
We divided our dataset into a set of training trajectories and evaluation trajectories. We then trained four latent-space models with the following varied input modalities: 

\begin{enumerate}
    \item RGB image only, as in \cite{kahn2021badgr} (Image)
    \item RGB image, heightmap and RGB map (Image + Maps)
    \item IMU, shock position and wheel RPM (Time-series)
    \item All Modalities (All)
\end{enumerate}

We trained each model using each loss function described in the previous section. We also implemented a baseline kinematic bicycle model (KBM) that leveraged the average wheel RPM to make predictions. Table \ref{tab:train_results} shows the accuracy of each model as the root mean squared error (RMSE) of the mean state prediction after 20 steps of forward-simulation. 
The model with the lowest score for a given loss (i.e.\ the best set of modalities) is bolded. The model with the lowest evaluation score for a given modality (i.e.\ the best loss function) is colored in red. 
Note that since the KBM is not a latent-space model, we copy its evaluation score across all loss columns.

\begin{table}[t]
    \centering
    \vspace*{0.05in}
    \begin{tabular}{c||c|c|c}
        & State & Reconstruction & Contrastive \\
        \hline
        KBM & 1.1638 & 1.1638 & 1.1638 \\
        Image &  0.5263 & \color{red}0.4740 & 0.4952\\
        Image + Maps & 0.3521 & \color{red}0.3386 & 0.3741\\
        Time Series & 0.2176 & 0.2285 & \color{red}0.1966 \\
        All & \textbf{0.1896} & \color{red}\textbf{0.1674} & \textbf{0.1958}\\
    \end{tabular}
    \caption{Model prediction results showing RMSE of mean state prediction for different models and loss functions.
    }
    \label{tab:train_results}
    \vspace*{-0.25in}
\end{table}

Overall, we can observe that adding additional modalities to the latent-space model results in improved prediction accuracy. The most noticeable improvement comes from adding top-down maps to the image-only model, yielding a 33\% decrease in prediction error. From our results, we can gather that the time-series data is very important to the overall dynamics prediction. This is evidenced by the large increase in model accuracy from adding the time-series data (roughly 45\% improvement from Image + Maps to All across all training procedures), and the relatively high accuracy of the time-series model. This is to be expected, as wheel RPM in particular is highly correlated with the velocity. However, we still observe that adding the image-based modalities to the time-series model yielded roughly a 15\% increase in model accuracy across all training procedures. We note that the performance of the time-series and all-modality models are essentially the same under the contrastive loss.

We also ran experiments to characterize the impact of exteroceptive sensing on out learned models in more challenging driving scenarios. In order to quantify this effect, we compared the prediction accuracy of time-series input only models to prediction accuracy of models incorporating exteroception from the maps and images in both the original dataset, and a new dataset separately collected exclusively in more uneven terrain. We define a trajectory difficulty metric as average change in height per second, which roughly corresponds to terrain steepness and unevenness. The original and new dataset had median trajectory difficulties of $0.0866 m/s$ and $0.2253 m/s$, respectively. 87\% of the trajectories in the new dataset were more difficult than the median difficulty in the original dataset.  Table~\ref{tab:hard_eval} shows the prediction accuracy of time-series and all-modality models on each evaluation set, as well as the percent improvement.

\begin{table}[t]
    \centering
    \vspace*{0.05in}
    \begin{tabular}{c||c|c|c}
         Dataset & Prop. Error & Prop. + Ext. Error & Improvement\\
         \hline
         Original & 0.2176 & 0.1896 & 13\%\\
         Difficult & 0.7313 & 0.5394 & 26\%\\
    \end{tabular}
    \caption{Comparison of Proprioception-only (Prop.) and Proptioception + Exteroception (Prop. + Ext.) Models on the Original and More Difficult Evaluation Datasets}
    \label{tab:hard_eval}
    \vspace*{-0.25in}
\end{table}

Overall, we observe that the additional vision-based modalities are more beneficial in the more challenging scenarios. This makes sense, as the difficulty of the terrain increases, it becomes increasingly difficult to accurately predict the future using proprioception alone.

\section{Conclusions and Future Work}
We have presented TartanDrive, a large-scale dataset for training of deep neural networks for dynamics prediction using multiple sensing modalities. We have also provided a benchmark of recent neural network architectures for dynamics prediction from high-dimensional inputs. We plan to continue collecting data and make the dataset potentially useful for tasks other than dynamics prediction, for instance imitation learning. 


An obvious direction for future work is the incorporation of these models into a navigation stack. While we have demonstrated acceptable dynamics prediction, it remains to be seen whether this improved dynamics prediction is sufficient for intelligent navigation over rough terrain. We believe that additional research will be necessary to create effective cost functions and planning algorithms that can also leverage the corpus of data we have collected for this work.

Another interesting direction for future work would be the combination of deep latent models and mapping-based approaches. As mentioned in the analysis section, the added partial observability of real-world navigation compared to simulation tasks renders auxiliary sensory prediction tasks largely unhelpful. Additionally, the latent models are not trained on sequences long enough to facilitate learning some form of SLAM. We believe that incorporating some sort of mapping-based approach could allow deep models to exhibit more long-term reasoning capabilities by storing past observations (and potentially features, like in Tung et al. \cite{tung2019learning}) on a map. 

{
\bibliographystyle{IEEEtran}
\bibliography{refs}

\begin{thebibliography}{10}
\providecommand{\url}[1]{#1}
\csname url@samestyle\endcsname
\providecommand{\newblock}{\relax}
\providecommand{\bibinfo}[2]{#2}
\providecommand{\BIBentrySTDinterwordspacing}{\spaceskip=0pt\relax}
\providecommand{\BIBentryALTinterwordstretchfactor}{4}
\providecommand{\BIBentryALTinterwordspacing}{\spaceskip=\fontdimen2\font plus
\BIBentryALTinterwordstretchfactor\fontdimen3\font minus
  \fontdimen4\font\relax}
\providecommand{\BIBforeignlanguage}[2]{{%
\expandafter\ifx\csname l@#1\endcsname\relax
\typeout{** WARNING: IEEEtran.bst: No hyphenation pattern has been}%
\typeout{** loaded for the language `#1'. Using the pattern for}%
\typeout{** the default language instead.}%
\else
\language=\csname l@#1\endcsname
\fi
#2}}
\providecommand{\BIBdecl}{\relax}
\BIBdecl

\bibitem{howard2007optimal}
T.~M. Howard and A.~Kelly, ``Optimal rough terrain trajectory generation for
  wheeled mobile robots,'' \emph{The International Journal of Robotics
  Research}, vol.~26, no.~2, pp. 141--166, 2007.

\bibitem{kahn2021badgr}
G.~Kahn, P.~Abbeel, and S.~Levine, ``Badgr: An autonomous self-supervised
  learning-based navigation system,'' \emph{IEEE Robotics and Automation
  Letters}, vol.~6, no.~2, pp. 1312--1319, 2021.

\bibitem{tremblay2020multimodal}
J.-F. Tremblay, T.~Manderson, A.~Noca \emph{et~al.}, ``Multimodal dynamics
  modeling for off-road autonomous vehicles,'' \emph{IEEE International
  Conference on Robotics and Automation}, 2020.

\bibitem{divconstrained}
S.~J. Wang, S.~Triest, W.~Wang \emph{et~al.}, ``Rough terrain navigation using
  divergence constrained model based reinforcement learning,'' in
  \emph{Conference on Robot Learning}.\hskip 1em plus 0.5em minus 0.4em\relax
  PMLR, 2021.

\bibitem{he2016identity}
K.~He, X.~Zhang, S.~Ren, and J.~Sun, ``Identity mappings in deep residual
  networks,'' in \emph{European conference on computer vision}.\hskip 1em plus
  0.5em minus 0.4em\relax Springer, 2016, pp. 630--645.

\bibitem{geyer2020a2d2}
\BIBentryALTinterwordspacing
J.~Geyer, Y.~Kassahun, M.~Mahmudi \emph{et~al.}, ``{A2D2: Audi Autonomous
  Driving Dataset},'' 2020. [Online]. Available: \url{https://www.a2d2.audi}
\BIBentrySTDinterwordspacing

\bibitem{waymo_open_dataset}
``Waymo open dataset: An autonomous driving dataset,'' 2019.

\bibitem{Argoverse}
M.-F. Chang, J.~W. Lambert, P.~Sangkloy \emph{et~al.}, ``Argoverse: 3d tracking
  and forecasting with rich maps,'' in \emph{Conference on Computer Vision and
  Pattern Recognition}, 2019.

\bibitem{Cordts2016Cityscapes}
M.~Cordts, M.~Omran, S.~Ramos \emph{et~al.}, ``The cityscapes dataset for
  semantic urban scene understanding,'' in \emph{Proc. of the IEEE Conference
  on Computer Vision and Pattern Recognition}, 2016.

\bibitem{RobotCarDatasetIJRR}
\BIBentryALTinterwordspacing
W.~Maddern, G.~Pascoe, C.~Linegar, and P.~Newman, ``{1 Year, 1000km: The Oxford
  RobotCar Dataset},'' \emph{The International Journal of Robotics Research},
  vol.~36, no.~1, pp. 3--15, 2017. [Online]. Available:
  \url{http://dx.doi.org/10.1177/0278364916679498}
\BIBentrySTDinterwordspacing

\bibitem{maturana2018real}
D.~Maturana, P.-W. Chou, M.~Uenoyama, and S.~Scherer, ``Real-time semantic
  mapping for autonomous off-road navigation,'' in \emph{Field and Service
  Robotics}.\hskip 1em plus 0.5em minus 0.4em\relax Springer, 2018, pp.
  335--350.

\bibitem{RUGD2019IROS}
M.~Wigness, S.~Eum, J.~G. Rogers \emph{et~al.}, ``A rugd dataset for autonomous
  navigation and visual perception in unstructured outdoor environments,'' in
  \emph{International Conference on Intelligent Robots and Systems}, 2019.

\bibitem{valada2016deep}
A.~Valada, G.~L. Oliveira, T.~Brox, and W.~Burgard, ``Deep multispectral
  semantic scene understanding of forested environments using multimodal
  fusion,'' in \emph{International symposium on experimental robotics}.\hskip
  1em plus 0.5em minus 0.4em\relax Springer, 2016, pp. 465--477.

\bibitem{dabbiru2020lidar}
L.~Dabbiru, C.~Goodin, N.~Scherrer, and D.~Carruth, ``Lidar data segmentation
  in off-road environment using convolutional neural networks (cnn),''
  \emph{SAE International Journal of Advances and Current Practices in
  Mobility}, vol.~2, no. 2020-01-0696, pp. 3288--3292, 2020.

\bibitem{jiang2020rellis3d}
P.~Jiang, P.~Osteen, M.~Wigness, and S.~Saripalli, ``Rellis-3d dataset: Data,
  benchmarks and analysis,'' 2020.

\bibitem{shaban2022semantic}
A.~Shaban, X.~Meng, J.~Lee \emph{et~al.}, ``Semantic terrain classification for
  off-road autonomous driving,'' in \emph{Conference on Robot Learning}.\hskip
  1em plus 0.5em minus 0.4em\relax PMLR, 2022, pp. 619--629.

\bibitem{battaglia2016interaction}
P.~W. Battaglia, R.~Pascanu, M.~Lai \emph{et~al.}, ``Interaction networks for
  learning about objects, relations and physics,'' in \emph{NIPS}, 2016.

\bibitem{lerer2016learning}
A.~Lerer, S.~Gross, and R.~Fergus, ``Learning physical intuition of block
  towers by example,'' in \emph{International conference on machine
  learning}.\hskip 1em plus 0.5em minus 0.4em\relax PMLR, 2016, pp. 430--438.

\bibitem{baradel2019cophy}
F.~Baradel, N.~Neverova, J.~Mille \emph{et~al.}, ``Cophy: Counterfactual
  learning of physical dynamics,'' in \emph{International Conference on
  Learning Representations}, 2019.

\bibitem{agrawal2016learning}
P.~Agrawal, A.~Nair, P.~Abbeel \emph{et~al.}, ``Learning to poke by poking:
  experiential learning of intuitive physics,'' in \emph{Proceedings of the
  30th International Conference on Neural Information Processing Systems},
  2016, pp. 5092--5100.

\bibitem{levine2018learning}
S.~Levine, P.~Pastor, A.~Krizhevsky \emph{et~al.}, ``Learning hand-eye
  coordination for robotic grasping with deep learning and large-scale data
  collection,'' \emph{The International Journal of Robotics Research}, vol.~37,
  no. 4-5, pp. 421--436, 2018.

\bibitem{finn2016unsupervised}
C.~Finn, I.~Goodfellow, and S.~Levine, ``Unsupervised learning for physical
  interaction through video prediction,'' \emph{Advances in neural information
  processing systems}, vol.~29, pp. 64--72, 2016.

\bibitem{tremblay2019automatic}
J.-F. Tremblay, M.~B{\'e}land, F.~Pomerleau \emph{et~al.}, ``Automatic 3d
  mapping for tree diameter measurements in inventory operations,''
  \emph{Journal of Field Robotics}, 2019.

\bibitem{hafner2019dream}
D.~Hafner, T.~Lillicrap, J.~Ba, and M.~Norouzi, ``Dream to control: Learning
  behaviors by latent imagination,'' \emph{International Conference on Learning
  Representations}, 2019.

\bibitem{hafner2019learning}
D.~Hafner, T.~Lillicrap, I.~Fischer \emph{et~al.}, ``Learning latent dynamics
  for planning from pixels,'' in \emph{International Conference on Machine
  Learning}.\hskip 1em plus 0.5em minus 0.4em\relax PMLR, 2019, pp. 2555--2565.

\bibitem{hafner2020mastering}
D.~Hafner, T.~Lillicrap, M.~Norouzi, and J.~Ba, ``Mastering atari with discrete
  world models,'' \emph{International Conference on Learning Representations},
  2020.

\bibitem{pezzementi2018comparing}
Z.~Pezzementi, T.~Tabor, P.~Hu \emph{et~al.}, ``Comparing apples and oranges:
  Off-road pedestrian detection on the national robotics engineering center
  agricultural person-detection dataset,'' \emph{Journal of Field Robotics},
  vol.~35, no.~4, pp. 545--563, 2018.

\bibitem{gresenzoff}
G.~Gresenz, J.~White, and D.~C. Schmidt, ``An off-road terrain dataset
  including images labeled with measures of terrain roughness.''

\bibitem{sivaprakasm2021}
M.~Sivaprakasam, S.~Triest, W.~Wang \emph{et~al.}, ``Improving off-road
  planning techniques with learned costs from physical interactions,'' in
  \emph{Proceedings - IEEE International Conference on Robotics and
  Automation}, Xi'an, China, May 2021.

\bibitem{Mai-2020}
J.~Mai, ``System design, modelling, and control for an off-road autonomous
  ground vehicle,'' Master's thesis, Carnegie Mellon University, Pittsburgh,
  PA, July 2020.

\bibitem{chang2018pyramid}
J.-R. Chang and Y.-S. Chen, ``Pyramid stereo matching network,'' in
  \emph{Proceedings of the IEEE Conference on Computer Vision and Pattern
  Recognition}, 2018, pp. 5410--5418.

\bibitem{wang2020tartanvo}
W.~Wang, Y.~Hu, and S.~Scherer, ``Tartanvo: A generalizable learning-based
  vo,'' \emph{Conference on Robot Learning}, 2020.

\bibitem{JMLR:v21:20-091}
\BIBentryALTinterwordspacing
R.~Tavenard, J.~Faouzi, G.~Vandewiele \emph{et~al.}, ``Tslearn, a machine
  learning toolkit for time series data,'' \emph{Journal of Machine Learning
  Research}, vol.~21, no. 118, pp. 1--6, 2020. [Online]. Available:
  \url{http://jmlr.org/papers/v21/20-091.html}
\BIBentrySTDinterwordspacing

\bibitem{muller2007dynamic}
M.~M{\"u}ller, ``Dynamic time warping,'' \emph{Information retrieval for music
  and motion}, pp. 69--84, 2007.

\bibitem{van2008visualizing}
L.~Van~der Maaten and G.~Hinton, ``Visualizing data using t-sne.''
  \emph{Journal of machine learning research}, vol.~9, no.~11, 2008.

\bibitem{oord2016wavenet}
A.~v.~d. Oord, S.~Dieleman, H.~Zen \emph{et~al.}, ``Wavenet: A generative model
  for raw audio,'' \emph{arXiv preprint arXiv:1609.03499}, 2016.

\bibitem{wu2018multimodal}
M.~Wu and N.~Goodman, ``Multimodal generative models for scalable
  weakly-supervised learning,'' \emph{Advances in Neural Information Processing
  Systems}, 2018.

\bibitem{hinton2002training}
G.~E. Hinton, ``Training products of experts by minimizing contrastive
  divergence,'' \emph{Neural computation}, vol.~14, no.~8, pp. 1771--1800,
  2002.

\bibitem{cao2014generalized}
Y.~Cao and D.~J. Fleet, ``Generalized product of experts for automatic and
  principled fusion of gaussian process predictions,'' \emph{arXiv preprint
  arXiv:1410.7827}, 2014.

\bibitem{cho2014learning}
K.~Cho, B.~Van~Merri{\"e}nboer, C.~Gulcehre \emph{et~al.}, ``Learning phrase
  representations using rnn encoder-decoder for statistical machine
  translation,'' \emph{Conference on Empirical Methods in Natural Language
  Processing}, 2014.

\bibitem{tung2019learning}
H.-Y.~F. Tung, R.~Cheng, and K.~Fragkiadaki, ``Learning spatial common sense
  with geometry-aware recurrent networks,'' in \emph{Proceedings of the
  IEEE/CVF Conference on Computer Vision and Pattern Recognition}, 2019, pp.
  2595--2603.

\end{thebibliography}


\begin{thebibliography}{1}

\bibitem{oord2016wavenet}
Aaron van~den Oord, Sander Dieleman, Heiga Zen, Karen Simonyan, Oriol Vinyals,
  Alex Graves, Nal Kalchbrenner, Andrew Senior, and Koray Kavukcuoglu.
\newblock Wavenet: A generative model for raw audio.
\newblock {\em arXiv preprint arXiv:1609.03499}, 2016.

\bibitem{zaheer2017deep}
Manzil Zaheer, Satwik Kottur, Siamak Ravanbakhsh, Barnabas Poczos, Ruslan
  Salakhutdinov, and Alexander Smola.
\newblock Deep sets.
\newblock {\em arXiv preprint arXiv:1703.06114}, 2017.

\bibitem{hinton2002training}
Geoffrey~E Hinton.
\newblock Training products of experts by minimizing contrastive divergence.
\newblock {\em Neural computation}, 14(8):1771--1800, 2002.

\bibitem{kingma2014adam}
Diederik~P Kingma and Jimmy Ba.
\newblock Adam: A method for stochastic optimization.
\newblock {\em arXiv preprint arXiv:1412.6980}, 2014.

\end{thebibliography}
}



\end{document}


\appendix

\subsection{Vehicle Frames and Parameters}

\begin{figure}
    \centering
    \includegraphics[scale=0.35]{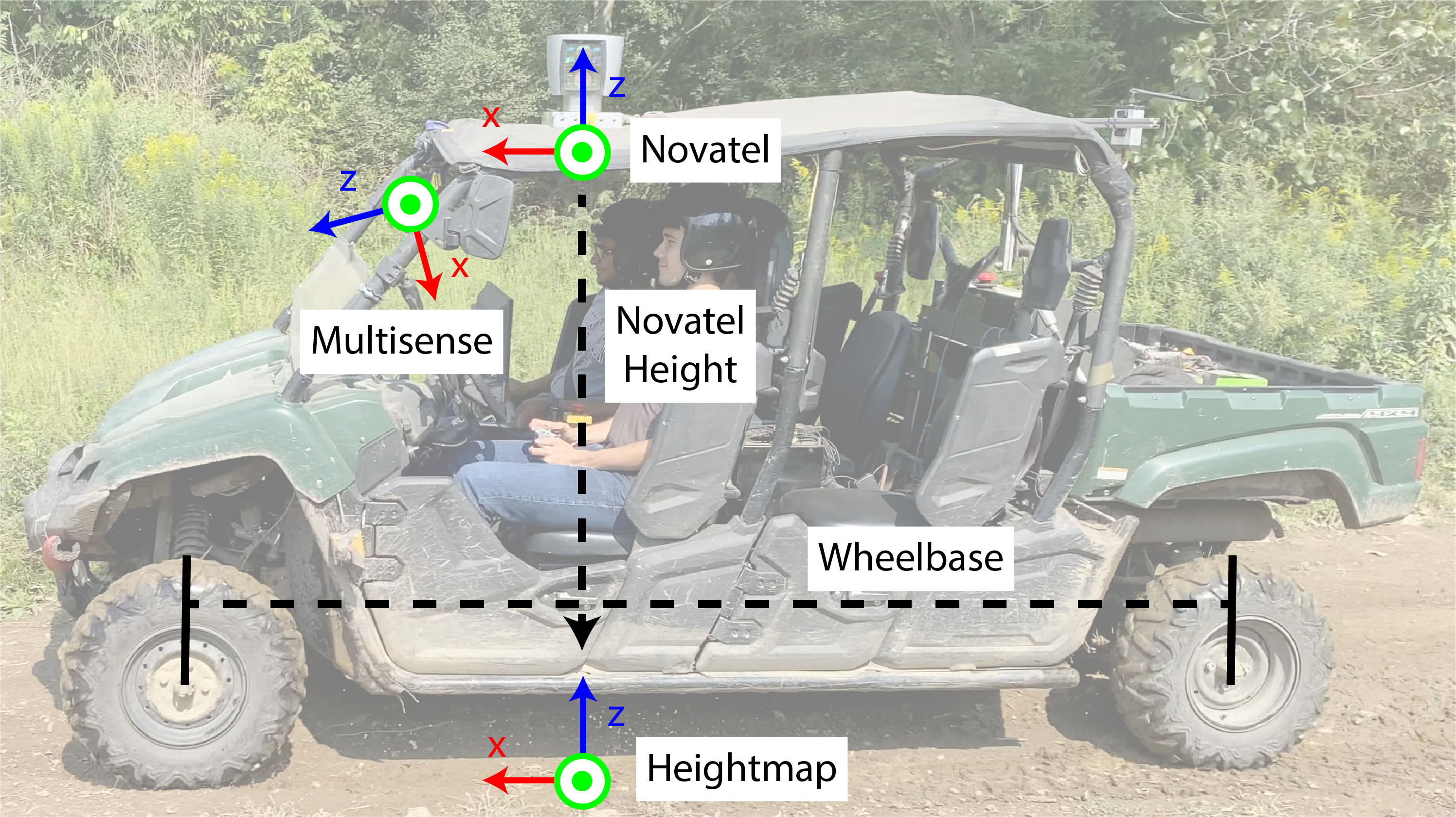}
    \caption{Qualitative description of the frames associated with the ATV}
    \label{fig:side_frames}
\end{figure}

There are three frames of note for the dataset: The novatel frame (which produces the state estimates), the Multisense frame (which produces the images and IMU), and the map frame (which produces the heightmap and RGB map). Their relative locations and orientations are provided in Figure \ref{fig:side_frames}. These quantities are also given explicitly in the dataset. We also provide the wheelbase and GPS height in Table \ref{tab:atv_params}.

\begin{table}[]
    \centering
    \begin{tabular}{c|c}
        Value & Quantity \\
        \hline
        Wheelbase & 3$m$ \\
        GPS height & 1.57$m$
    \end{tabular}
    \caption{Additional Geometrical Parameters for the ATV}
    \label{tab:atv_params}
\end{table}

\subsection{Network Architecture and Training Procedure}
In this section, we elaborate more on our network architectures and training procedures. The algorithm for generation state and observation predictions is presented in Algorithm \ref{algo:latent_model}. The general algorithm for encoding and decoding both image and time-series data is presented in Algorithms \ref{algo:upsample_block}-\ref{algo:cnn_decoder}. The temporal downsample block follows the implementation of WaveNet \cite{oord2016wavenet} (i.e. gated, dilated, causal convolutions). However, as there is no temporal order to the latent code, temporal upsampling is handled simply by 1D convolution and upsampling along the time dimension. We present the full list of neural network architectures in Tables \ref{tab:vcnn_encoder_arch}-\ref{tab:tcnn_decoder_arch}. We present our training hyperparameters in Table \ref{tab:train_hyperparams}. Since we evaluate multiple different loss types, we add an additional column denoting which experiments used which hyperparameters (with 'R' standing for reconstruction and 'C' for contrastive). 

\begin{algorithm*}
\DontPrintSemicolon
\caption{Latent Model Forward Pass}
\label{algo:latent_model}
\KwIn{Modality set $M$, initial state $x_0$, initial observations $\{o^m_0, \forall m \in M\}$, action sequence $a_{1:T}$, modality prediction set $\tilde{M}$. Encoders $e^m_\psi, \forall m \in M$, Decoders $d^m_\psi, \forall m \in \tilde{M}$, latent model $f_\theta(z, a)$, action encoder $g_\psi(a)$, state decoder $d^{state}_\psi$}
\KwOut{State predictions $\tilde{x_{1:T}}$, observation predictions $\{o^m_{1:t},  \forall m \in \tilde{M}\}$}

\For{$m \in M$}{
    $p^m(z) \leftarrow e^m_\psi(o^m_0)\hfill \triangleleft$ Encode each observation into $\mathcal{Z}$ \;
}

$z_0 =$ aggregate$(\{p^m(z), \forall m \in M\}) \hfill \triangleleft$ Use Deepsets \cite{zaheer2017deep} or Product of Experts \cite{hinton2002training} to get single $z$\;

\For{$t \in 1:T$}{
    $z_t = f_\theta(z,  g_\psi(a_{t-1}))\hfill \triangleleft$ Embed action and predict next latent state \;
    $x_t = d^{state}_\psi(z_t)\hfill \triangleleft$ Decode state\;
    \For{$m \in \tilde{M}$}{
        $o^m_{t+1} = d_\psi(z_t)\hfill \triangleleft$ Decode observation\;
    }
}

return $x_{1:T}, \{o^m_{1:t},  \forall m \in \tilde{M}\}$ \;
\end{algorithm*}

\begin{algorithm}
\DontPrintSemicolon
\caption{Upsample Block}
\label{algo:upsample_block}
\KwIn{Image input $x$, upsample factor $s$, convolution kernel $K$, activation function $f$}
\KwOut{Upsampled image output $\tilde{x}$}

$x \leftarrow $ linear interpolate$(x, s)$ \;
$x \leftarrow x * K$ \;
$x \leftarrow f(x)$ \;
return $x$ \;
\end{algorithm}

\begin{algorithm}
\DontPrintSemicolon
\caption{Downsample Block}
\label{algo:downsample_block}
\KwIn{Image input $x$, downsample factor $s$, convolution kernel $K$, activation function $f$}
\KwOut{Downsampled image output $\tilde{x}$}

$x \leftarrow x * K$ \;
$x \leftarrow f(x)$ \;
$x \leftarrow $ linear interpolate$(x, s)$ \;
return $x$ \;
\end{algorithm}

\begin{algorithm}
\DontPrintSemicolon
\caption{CNN Encoder}
\label{algo:cnn_encoder}
\KwIn{Image input $x$, downsample blocks $D_\psi$, MLP $f_\theta$}
\KwOut{Latent distribution $p(z)$}

\For{$d_\psi$ in D}{
    $x \leftarrow d_\psi(x)\hfill \triangleleft$ using Algorithm \ref{algo:downsample_block} or \cite{oord2016wavenet}\;
}
$x \leftarrow$ flatten$(x) \hfill \triangleleft$ Flatten $x$ to 1D \;
$\mu, \sigma \leftarrow f_\theta(x)$ \;
return $\mathcal{N}(\mu, \sigma)$ \;

\end{algorithm}

\begin{algorithm}
\DontPrintSemicolon
\caption{CNN Decoder}
\label{algo:cnn_decoder}
\KwIn{Latent vector $z$, upsample blocks $U_\psi$, MLP $f_\theta$}
\KwOut{Image reconstruction $\tilde{X}$}

$x \leftarrow f_\theta(x)$ \;
$x \leftarrow$ pad\_front $(x, 2) \hfill \triangleleft$ $x \in \{1 \times 1 \times |x|\}$ \;

\For{$u_\psi$ in U}{
    $x \leftarrow u_\psi(x) \hfill \triangleleft$ using Algorithm \ref{algo:upsample_block}\;
}
return $x$ \;

\end{algorithm}

\begin{table}[]
    \centering
    \begin{tabular}{c||c|c|c|c}
        Layer & Input Dim & Output Dim & Kernel Size & Activation \\
        \hline
        Downsample 1 & $3 \times 128 \times 128$ & $4 \times 64 \times 64$ & $3 \times 3$ & ReLU \\
        Downsample 2 & $4 \times 64 \times 64$ & $8 \times 32 \times 32$ & $3 \times 3$ & ReLU \\
        Downsample 3 & $8 \times 32 \times 32$ & $16 \times 16 \times 16$ & $3 \times 3$ & ReLU \\
        Downsample 4 & $16 \times 16 \times 16$ & $32 \times 8 \times 8$ & $3 \times 3$ & ReLU \\
        Flatten & $32 \times 8 \times 8$ & $2048$ & - & - \\
        MLP & $2048$ & $2 \times |\mathcal{Z}|$ & - & Tanh \\
        Gaussian & $2 \times |\mathcal{Z}|$ & $\mathcal{N} \in \mathcal{Z}$ & - & - \;
    \end{tabular}
    \caption{Visual CNN Encoder Architecture}
    \label{tab:vcnn_encoder_arch}
\end{table}

\begin{table}[]
    \centering
    \begin{tabular}{c||c|c|c|c}
        Layer & Input Dim & Output Dim & Kernel Size & Activation \\
        \hline
        Downsample 1 & $\{1, 3\} \times 64 \times 64$ & $4 \times 32 \times 32$ & $3 \times 3$ & ReLU \\
        Downsample 2 & $4 \times 32 \times 32$ & $8 \times 16 \times 16$ & $3 \times 3$ & ReLU \\
        Downsample 3 & $8 \times 16 \times 16$ & $16 \times 8 \times 8$ & $3 \times 3$ & ReLU \\
        Downsample 4 & $16 \times 8 \times 8$ & $32 \times 4 \times 4$ & $3 \times 3$ & ReLU \\
        Flatten & $32 \times 4 \times 4$ & $512$ & - & - \\
        MLP & $512$ & $2 \times |\mathcal{Z}|$ & - & Tanh \\
        Gaussian & $2 \times |\mathcal{Z}|$ & $\mathcal{N} \in \mathcal{Z}$ & - & - \;
    \end{tabular}
    \caption{Local Map CNN Encoder Architecture}
    \label{tab:hcnn_encoder_arch}
\end{table}

\begin{table}[]
    \centering
    \begin{tabular}{c||c|c|c|c|c}
        Layer & Input Dim & Output Dim & Size & Dilation & Activation \\
        \hline
        Downsample 1 & $\{4, 9\} \times 20$ & $\{4, 9\} \times 20$ & 2 & 2 & \cite{oord2016wavenet} \\
        Downsample 2 & $\{4, 9\} \times 20$ & $\{4, 9\} \times 20$ & 2 & 4 & \cite{oord2016wavenet} \\
        Downsample 3 & $\{4, 9\} \times 20$ & $\{4, 9\} \times 20$ & 2 & 8 & \cite{oord2016wavenet} \\
        Downsample 4 & $\{4, 9\} \times 20$ & $\{4, 9\} \times 20$ & 2 & 16 & \cite{oord2016wavenet} \\
        Flatten & $\{4, 9\} \times 20$ & $\{80, 180\}$ & - & - & - \\
        MLP & $\{80, 180\}$ & $2 \times |\mathcal{Z}|$ & - & - & Tanh \\
        Gaussian & $2 \times |\mathcal{Z}|$ & $\mathcal{N} \in \mathcal{Z}$ & - & - \;
    \end{tabular}
    \caption{Temporal CNN Encoder Architecture}
    \label{tab:tcnn_encoder_arch}
\end{table}

\begin{table}[]
    \centering
    \begin{tabular}{c||c|c|c|c}
        Layer & Input Dim & Output Dim & Kernel Size & Activation \\
        \hline
        MLP & $|\mathcal{Z}|$ & $128$ & - & Tanh \\
        Unflatten & $128$ & $128 \times 1 \times 1$ & - & - \\
        Upsample 1 & $128 \times 1 \times 1$ & $32 \times 4 \times 4$ & $3 \times 3$ & ReLU \\
        Upsample 2 & $32 \times 4 \times 4$ & $16 \times 8 \times 8$ & $3 \times 3$ & ReLU \\
        Upsample 3 & $16 \times 8 \times 8$ & $8 \times 16 \times 16$ & $3 \times 3$ & ReLU \\
        Upsample 4 & $8 \times 16 \times 16$ & $4 \times 32 \times 32$ & $3 \times 3$ & ReLU \\
        Upsample 5 & $4 \times 32 \times 32$ & $3 \times 128 \times 128$ & $3 \times 3$ & ReLU \\
    \end{tabular}
    \caption{Visual CNN Decoder Architecture}
    \label{tab:vcnn_decoder_arch}
\end{table}

\begin{table}[]
    \centering
    \begin{tabular}{c||c|c|c|c}
        Layer & Input Dim & Output Dim & Kernel Size & Activation \\
        \hline
        MLP & $|\mathcal{Z}|$ & $128$ & - & Tanh \\
        Unflatten & $128$ & $128 \times 1 \times 1$ & - & - \\
        Upsample 1 & $128 \times 1 \times 1$ & $32 \times 4 \times 4$ & $3 \times 3$ & ReLU \\
        Upsample 2 & $32 \times 4 \times 4$ & $16 \times 8 \times 8$ & $3 \times 3$ & ReLU \\
        Upsample 3 & $16 \times 8 \times 8$ & $8 \times 16 \times 16$ & $3 \times 3$ & ReLU \\
        Upsample 4 & $8 \times 16 \times 16$ & $4 \times 32 \times 32$ & $3 \times 3$ & ReLU \\
        Upsample 5 & $4 \times 32 \times 32$ & $3 \times 64 \times 64$ & $3 \times 3$ & ReLU \\
    \end{tabular}
    \caption{Local Map CNN Decoder Architecture}
    \label{tab:hcnn_decoder_arch}
\end{table}

\begin{table}[]
    \centering
    \begin{tabular}{c||c|c|c|c|c}
        Layer & Input Dim & Output Dim & Kernel Size & Activation \\
        \hline
        Unflatten & $|\mathcal{Z}|$ & $1 \times |\mathcal{Z}|$ & - & - \\
        Upsample 1 & $1 \times |\mathcal{Z}|$ & $2 \times 64$ & 2 & Tanh \\
        Upsample 1 & $2 \times 64$ & $4 \times 32$ & 2 & Tanh \\
        Upsample 1 & $4 \times 32$ & $8 \times 16$ & 2 & Tanh \\
        Upsample 1 & $8 \times 16$ & $16 \times 8$ & 2 & Tanh \\
        Upsample 1 & $16 \times 8$ & $20 \times \{4, 9\}$ & 2 & Tanh \\
    \end{tabular}
    \caption{Temporal CNN Decoder Architecture}
    \label{tab:tcnn_decoder_arch}
\end{table}

\begin{table}[]
    \centering
    \begin{tabular}{c||c|c|c|c|c}
        Layer & Input Dim & Output Dim & Activation \\
        \hline
        Action Encode 1 & $2$ & $16$ & Tanh \\
        Action Encode 2 & $16$ & $16$ & Tanh \\
        GRU & $(128, 23)$ & $128, 128$ & - \\
        State Decode 1 & $128$ & $128$ & Tanh \\
        State Decode 2 & $128$ & $\mathcal{N} \in \mathbb{R}^7$ & - \\
    \end{tabular}
    \caption{Latent Model Architecture}
    \label{tab:latent_model_hyperparams}
\end{table}

\begin{table}[]
    \centering
    \begin{tabular}{c||c|c}
        Hyperparameter & Value & Experiment \\
        \hline
        Optimizer & Adam \cite{kingma2014adam} & All \\
        Learning Rate & $1e-3$ & All \\
        Epochs & 5000 & All \\
        Batch Size & 64 & All \\
        Gradient Steps Per Epoch & 10 & All \\
        Gradient Norm Clip & 100.0 & All \\
        Train Timesteps & 20 & All \\
        RGB Image Loss Scale & 100 & R \\
        RGB Map Loss Scale & 100 & R \\
        Heightmap Loss Scale & 1 & R \\
        IMU Loss Scale & 0.1 & R \\
        Wheel RPM Loss Scale & 0.1 & R \\
        Contrastive Scale & 10.0 & C \\
        EMA $\tau$ & 0.05 & C \\
    \end{tabular}
    \caption{Training Hyperparameters}
    \label{tab:train_hyperparams}
\end{table}

\subsection{Algorithm for T-SNE Clustering}
In this section, we describe in more detail our algorithm for performing time-series clustering. This is presented in Algorithm \ref{algo:tsne}.

\begin{algorithm*}
\DontPrintSemicolon
\caption{T-SNE Clustering}
\label{algo:tsne}
\KwIn{Dataset $\mathcal{D}$ (binned by velocity), consisting of states $s_{1:T}$ and actions $a_{1:T}$, time window $k$, numbers of clusters $n$, }
\KwOut{Cluster mappings $c_{1:T}$ and t-SNE embeddings $z_{1:T}$ for each timestep}

$f_t = flatten(a_{t:t+k}), \forall t$ \hfill $\triangleleft$ Get features for each state by flattening actions over the window\;
$c_{1:k} = kmeans(f_{1:T})$ \hfill $\triangleleft$ Perform k-means to get cluster centers\;
$\mathcal{T}_t = (s_t)^{-1}, \forall t$ \hfill $\triangleleft$ Compute the transform to start all state differences at 0,0\;
$\Delta s_{1:T} = \mathcal{T}_t(s_{t+k} - s_{t}), \forall t$ \hfill $\triangleleft$ Compute state differences for all states\;
$z_{1:T} = tsne(\Delta s_{1:T})$ \hfill $\triangleleft$ Perform t-SNE on the state differences\;
\end{algorithm*}

\subsection{T-SNE figures For Dynamical Variation Experiment}
\begin{figure*}[]
    \centering
    \label{fig:tsne_all}
    \includegraphics[scale=0.55]{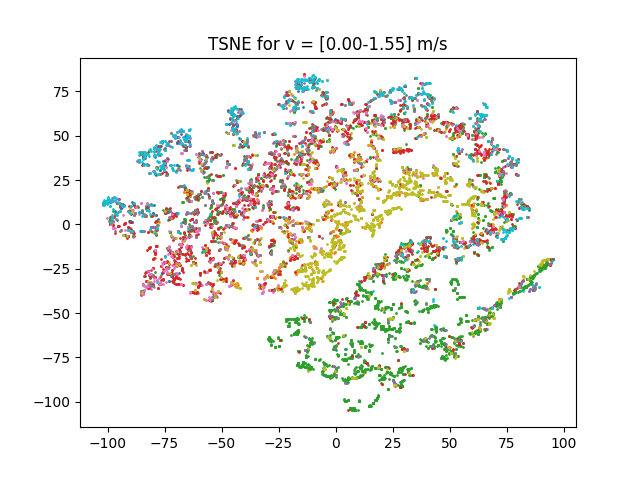}
    \includegraphics[scale=0.55]{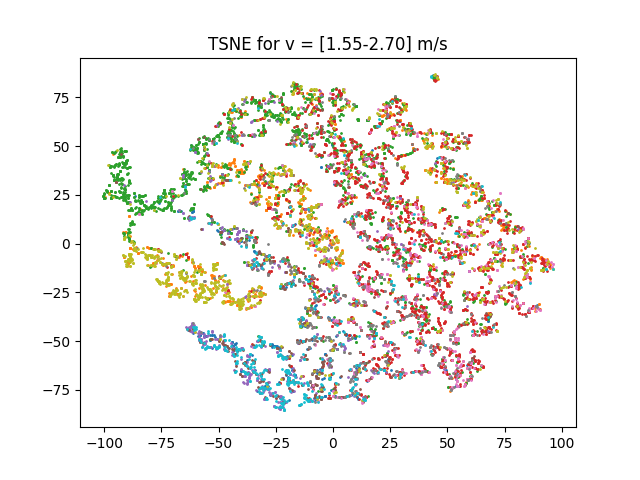}
    \includegraphics[scale=0.55]{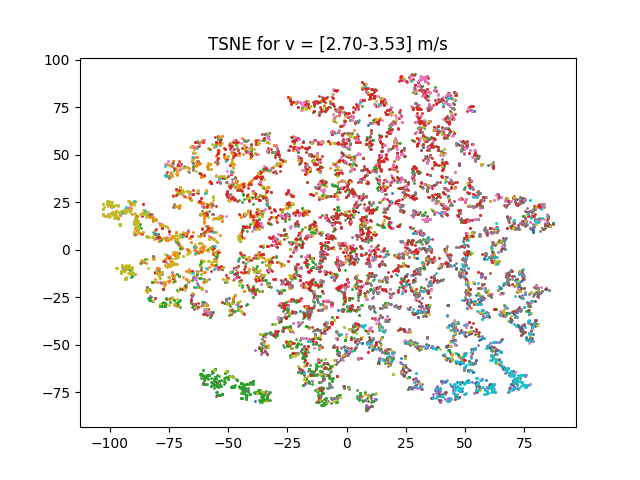}
    \includegraphics[scale=0.55]{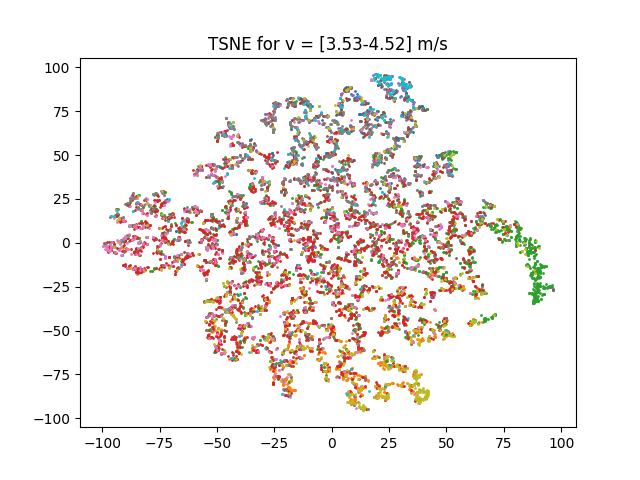}
    \includegraphics[scale=0.55]{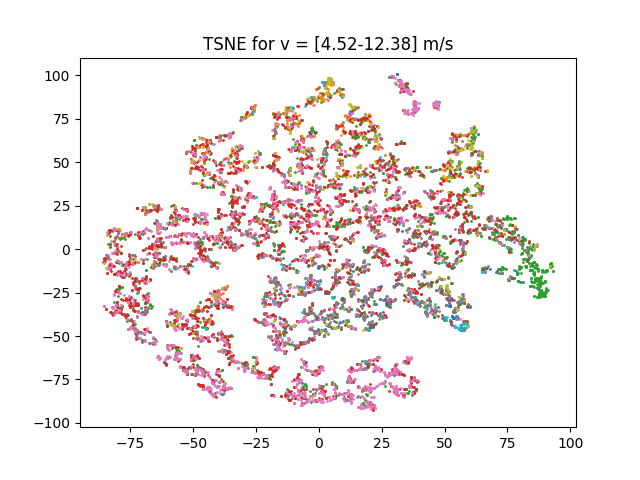}
    \caption{The t-SNE visualizations for all five velocity bins.}
    \label{fig:tsne_plots}
\end{figure*}

\begin{figure*}[h]
    \centering
    \includegraphics[scale=0.4]{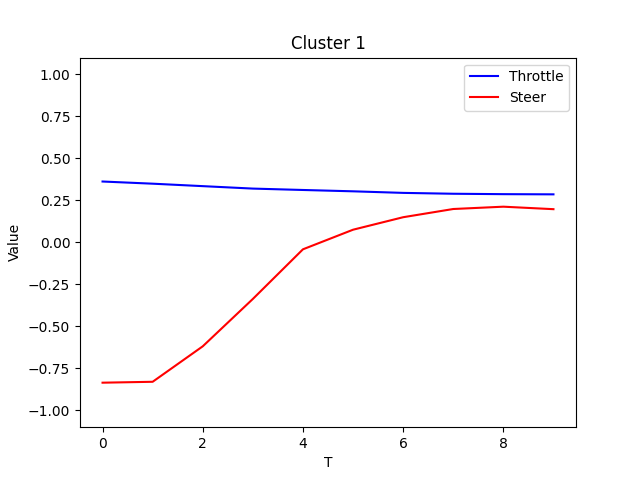}
    \includegraphics[scale=0.4]{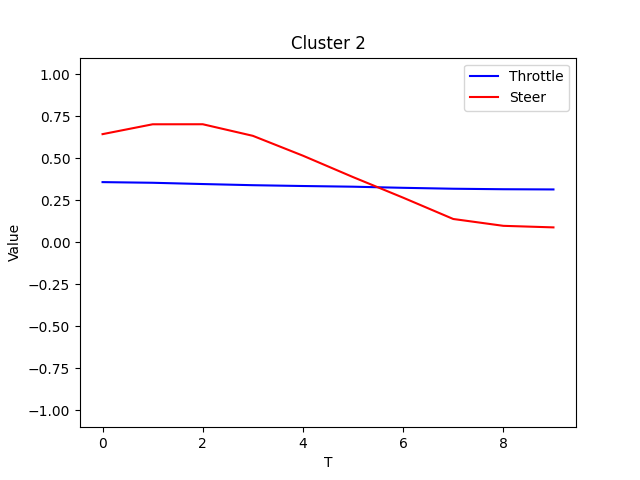}
    \includegraphics[scale=0.4]{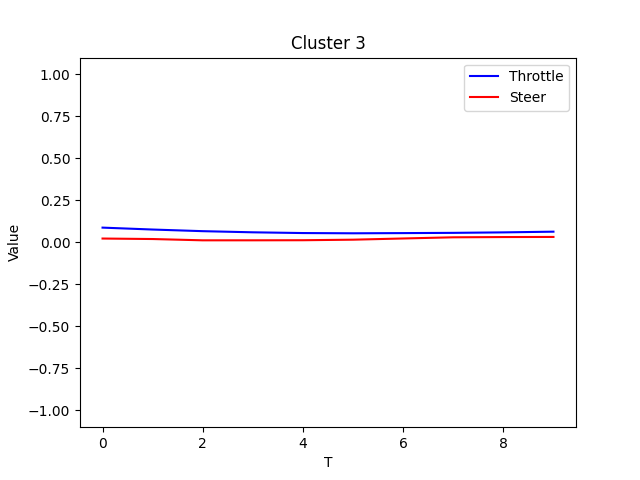}
    \includegraphics[scale=0.4]{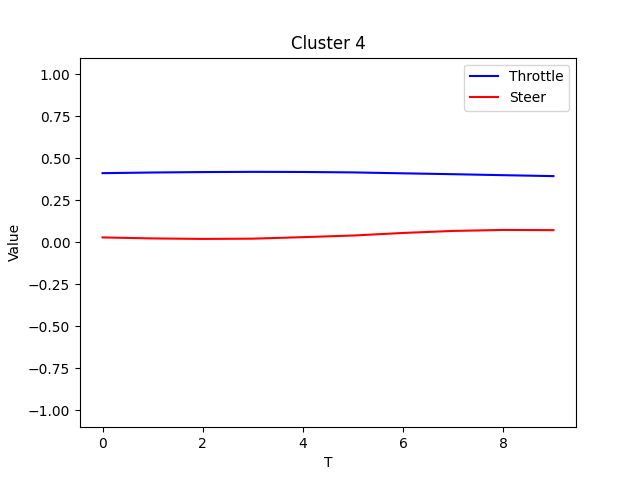}
    \includegraphics[scale=0.4]{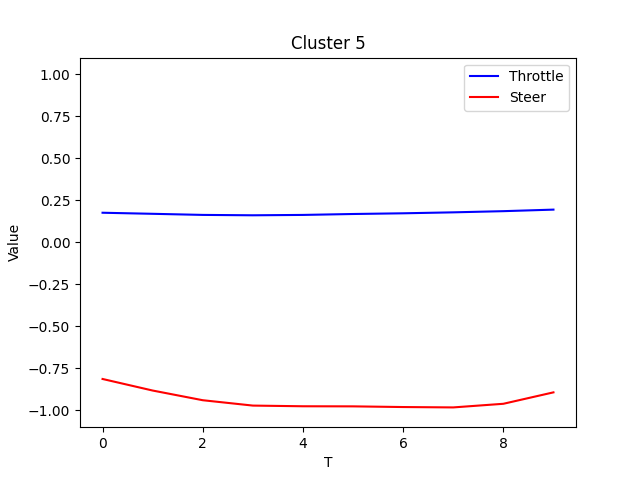}
    \includegraphics[scale=0.4]{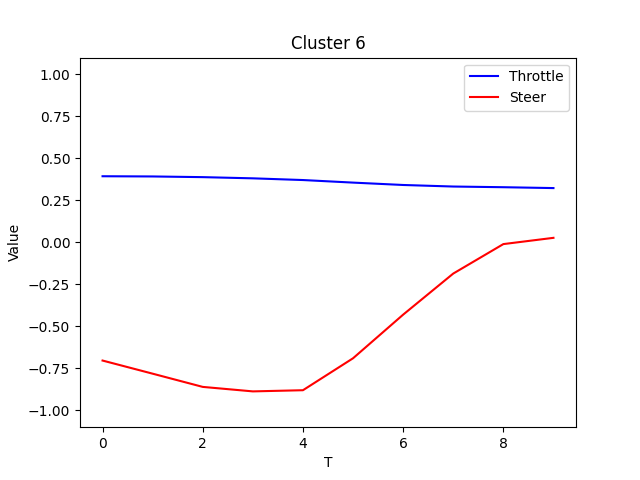}
    \includegraphics[scale=0.4]{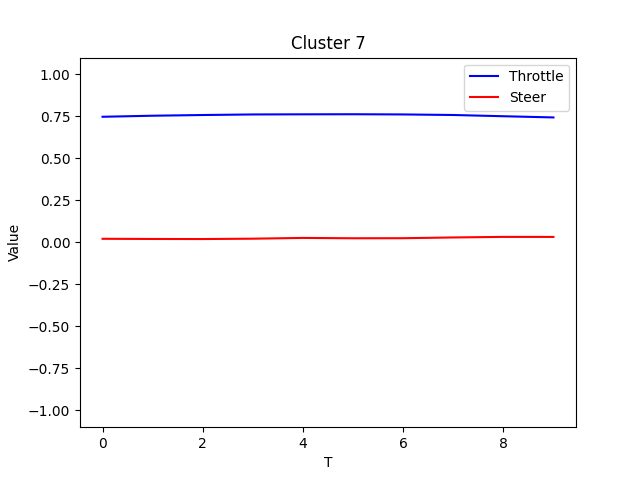}
    \includegraphics[scale=0.4]{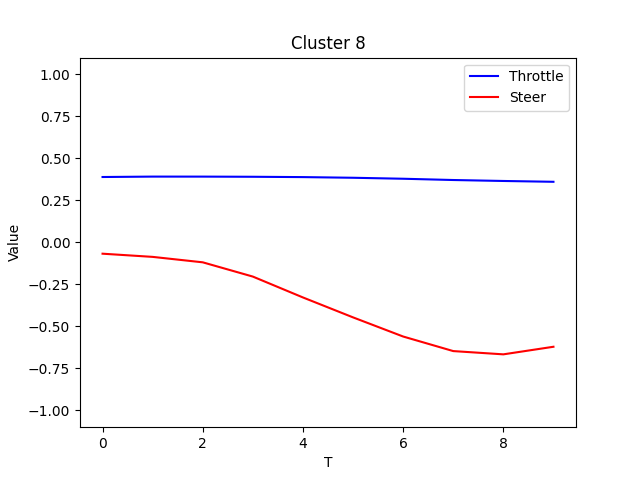}
    \includegraphics[scale=0.4]{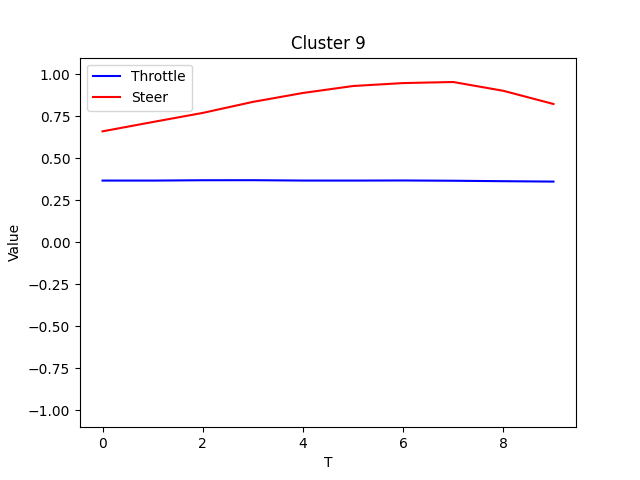}
    \includegraphics[scale=0.4]{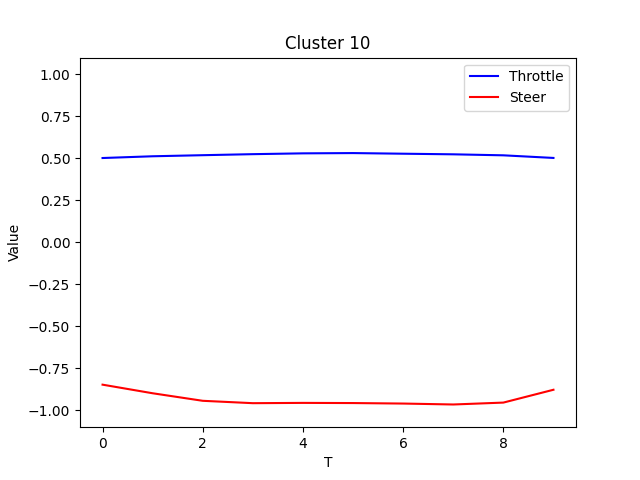}
    \caption{The cluster centroids for the motivational experiment}
    \label{fig:action_clusters}
\end{figure*}

The full set of t-SNE figures and clusters from our motivational experiment are provided in Figures \ref{fig:tsne_plots} and \ref{fig:action_clusters}, respectively. The hyperparameters for the experiment are provided in Table \ref{tab:tsne_hyperparams}.

\begin{table}[]
    \centering
    \begin{tabular}{c|c}
         Hyperparameter & Value \\
         \hline
         \# Subsequences & 10000 \\
         Sequence length & 10 \\
         \# Clusters & 10 \\
         \# Velocity Bins & 5 \\
         Clustering Distance Metric & Euclidean \\
    \end{tabular}
    \caption{Motivational Experiment Hyperparameters}
    \label{tab:tsne_hyperparams}
\end{table}

{
\bibliography{refs}
\bibliographystyle{unsrt}
}
